\documentclass[aps,prl,reprint,groupedaddress,showkeys]{revtex4-2}

\usepackage{braket}
\usepackage{amsmath}
\usepackage{graphicx}

\usepackage{CJKutf8}

\usepackage{array}
\usepackage{booktabs}
\usepackage{caption}
\usepackage{multirow}

\usepackage{threeparttable}
\usepackage[colorlinks=true, linkcolor=blue, urlcolor=blue, citecolor=blue]{hyperref}
\hypersetup
{
	colorlinks=true,
	linkcolor=blue,
	filecolor=blue,
	urlcolor=blue,
	citecolor=cyan,
}

\begin{document}


\title{Universal language model with the intervention of quantum theory}

\author{D.-F. Qin}
\email{df_ala@outlook.com}
\affiliation{ School of Physics and Electronic Science, East China Normal University, Shanghai, China}


\date{\today}

\begin{abstract}

This paper examines language modeling based on the theory of quantum mechanics. It focuses on the introduction of quantum mechanics into the symbol-meaning pairs of language in order to build a representation model of natural language. At the same time, it is realized that word embedding, which is widely used as a basic technique for statistical language modeling, can be explained and improved by the mathematical framework of quantum mechanics. On this basis, this paper continues to try to use quantum statistics and other related theories to study the mathematical representation, natural evolution and statistical properties of natural language. It is also assumed that the source of such quantum properties is the physicality of information. The feasibility of using quantum theory to model natural language is pointed out through the construction of a experimental code. The paper discusses, in terms of applications, the possible help of the theory in constructing generative models that are popular nowadays. A preliminary discussion of future applications of the theory to quantum computers is also presented.
\end{abstract}


\maketitle

\section{I.Introduction}
We assume that natural language has a property analogous to quantum mechanics,  and the rationale for adopting this analogy comes not only from noticing that the meanings of natural language can be analogous described by in the form of superposition of states, but also from the fact that we notice that the representation of natural language symbols has a duality in which the symbols-meanings correspond to each other.

At the same time, natural language processing (NLP) methods based on statistical implementations have been popular for decades and continue to evolve.It makes us wonder whether we can go further than statistical theory and consider introducing the relevant theories of quantum mechanics into the modeling of natural language. Meanwhile, in the last decade or so, NLP has generally adopted the approach of converting natural language symbols into some kind of mathematical representation for processing, and this approach, known as word embedding, has achieved surprisingly good performance in universal language processing tasks.We see that the technical idea of completely converting the symbols that make up natural language into a numerical representation for processing to build a universal language model(ULM) is highly feasible and possible. The experimental progress fed back from these two real-world applications motivates us to explore building natural language models based on quantum theory.

Within the past century, quantum theory has become more and more perfect as mankind has studied the physical world in depth.  There is a general ambiguity of meaning in human communication using natural language. This has inspired a group of researchers to try to give an explanation to this phenomenon through the equally ``elusive''nature of quantum mechanics\cite{sentence-sim}. We should note that logically rigorous quantum mechanics can provide us with formal mathematical tools for modeling a usable set of quantum languages(QLM).

We bring together a number of presuppositions that provide the basis for the modeling. These assumptions are obtained by observing humans' own use of natural language over time and remain to be proven.

\textbf{Hypothesis I: There is a symbol-meaning duality in natural language. At the same time meaning exhibits a superposition of underlying concepts.}The set of characters of a natural language describes the simultaneous existence of many different meanings shaped like superposition states. We assume that natural language symbols are referents of some quantum system.

\textbf{Hypothesis II: Natural language is recursively composed of multiple levels of subsystems that interact with each other when combined.}This interaction causes a change in the state of the quantum system to which the symbols stand in for, allowing the symbols to exhibit specific meanings in a specific linguistic context.

\textbf{Hypothesis III: According to quantum theory, the state function of a quantum system provides information that contains all the properties of the system.}This is a more basic assumption taken based on experimental feedback.

In the first half of this paper, the full model based on these assumptions seems to explain many of the existing approaches to language modeling fairly reasonably well. And it gives us the ability to make some speculations about the properties of natural language through the model in conjunction with quantum theory.

This paper further explores, for example, the key steps in constructing a universal natural language model and points out that the use of quantum states to represent sequences of natural language symbols and can naturally contain the effect of sequential contextual information on words. In a further attempt, this paper uses quantum statistics to study the macroscopic properties of a large number of symbolically referred quantum systems when they are aggregated.
 
This leads us to a more fundamental conjecture to explain: ``Why can natural language be modeled in this quantum-theoretic form?''to make more essential speculations:

\textbf{Physicality of information: information exists on some physical structure.} Obviously, this physical structure can be described not only by classical mechanics or classical statistical theory, but also by quantum theory. This makes it possible to apply a rigorous quantum mechanical theory to describe the status, evolution and statistical properties of natural language.

Similar realization is also mentioned in Randall's study\cite{Landauer}. This paper attempts to investigate such ideas deeper into QLM.

It is worth mentioning that during the construction of the model, we found relevant models that make connections with the current NLP base technology of word embedding. On the one hand, the existing models should characterize the feasibility of our modeling ideas, and on the other hand, the models provide theoretical explanations for mainstream methods. In turn, the mainstream models can be complemented from the principles.

In terms of application, between we still don't have practical manipulation and measurement of the complete quantum state related technology. In this paper, we still try to do some validation for the model on the basis of neural network. We can use the principle of neural network can simulate the process of human learning language. Rather than discussing the building of specific neural networks in depth as in computer science, this paper uses neural networks as a tool to build algorithms and avoid as much as possible the influence of engineering details on the model.

In the context of treating language as a quantum statistical system, we validate the evolution of the quantum statistical properties of language over time on a simple temporal-containing corpus and correlate the results with social phenomena.

In summary, the conclusions of this paper show a facilitating role in four aspects: providing an explanation for word embedding techniques in NLP processing, as well as new training ideas for machine learning's, attempting to extend the research in information physics to the quantum domain, and finally offering the possibility of using quantum statistical methods in Mathematical Linguistics.

\section{II.Related Work}
The field of natural language processing is in another period of rapid development. The Large Language Models represented by GPT, Deepseek and so on have shown landmark performance.

We first take a quick look at the development of mainstream models in the field of natural language, and then focus on the research related to the immature concept of quantum language modeling.
\subsection{2.1 Mainstream Approaches to NLP}

Looking back at the past of the natural language processing field, along with the birth of the NLP problem in the field of computers and artificial intelligence, there have been many different implementations in the past seventy years or so. We list here some of the more recognized development nodes.

In the last century, early ideas in the field of NLP were based on manually written language processing rules including conceptual dependency theory, expert system and other methods.Beginning in the nineties, methods such as Hidden Markov Models, Conditional Random Fields, etc. made statistical tools step into the researchers' view. By the time N-gram models were widely studied, statistical methods had become an important tool for machine learning.

In the last two decades, with the improvement of neural network as a tool, natural language processing architectures such as RNN, LSTM, and GRU have been unfolded under the concept of deep learning. After the transition of two important models, word2vec\cite{word2vec} and seq2seq\cite{seq2seq}, the Transformer model architecture, which introduces the Encoder-Decoder structure and Self-Attention mechanism \cite{attention2017}, replaces the Recurrent Neural Network (RNN) to become the new dominant model of NLP.On the transformer model mechanism using Pre-training mechanism a large number of semi-supervised learning Gpt series models \cite{gpt1} \cite{gpt2} \cite{gpt3} \cite{gpt4} , which is worth mentioning individually is the introduction of human feedback in the training process of the RLHF mechanism \cite{ RLHF} of InstructGPT\cite{igpt} .

The subsequent overall development of the field of Large Language Modeling has shown a trend of intertwined commercial incentives and scientific research prompting rapid iteration of models. Each NLP model can be equated to a highly task-specific neural network architecture, its accompanying training methodology, and, if necessary, included the dataset used for training.

Current cutting-edge issues of interest in the field include multimodality enabling LLMs to process data such as images, sound, video, etc. across NLP boundaries. Methods for fine-tuning parameters enable model performance improvement, model interpretability, and other issues.The advent of inference modeling has allowed LLM to demonstrate some machine thinking capabilities.

\subsection{2.2 An Attempt on QLM}
It may be noted that such constant successive iterations have led to the emergence of new models that carry some of the techniques of the previous models.If we compare the whole lineage of technological development to a large tree, we can say that the trunk of this tree the NLP's current dominant idea is based on language models of a statistical nature.It is clear that statistical methods have achieved amazing results, and it is equally important to explore possible theoretical models on other branches. Such thoughts lead us to look at the already well-established quantum theory.

Under the proposition of quantum language modeling, the keywords of early research point to the problem of quantum IR. Although the problems studied in IR are slightly different from the original intent of this paper.IR is concerned with retrieving needed information from unstructured data. The core is matching documents and queries. The field of NLP is dedicated to enabling computers to understand, interpret and generate human language.

Under this field, it is important to mention the results of Sordoni et al.\cite{QLM}. Inspired by the classical N-gram model, where words are treated as quantum events, this study introduces quantum probability into statistical language modeling.  A logically complete set of design architectures was provided for the still unclear concept of QLM. Later work by Zhang et al\cite{cnnqlm2022} . generalized Sordoni's model by improving it to be compatible with complex neural networks to replace the slightly outdated maximum fitness estimation for estimating parameters. Of interest is the inclusion of a mechanism in Zhang's model that uses complex phases to encode word order. Doing so gives practical significance to the unused complex imaginary part of Sordoni's model.The underlying technical idea of these models is based on bag-of-words models with the goal of generalizing earlier statistical language models using quantum statistical theory.

IR problems and NLP tasks both have some common technical foundations. For example, the language modeling that is the focus of this paper is a fundamental concept common to both fields, and the concern lies in how to mathematically model natural language. Based on this connection, we are very pleased to list related work as referenced results.

There are also scattered attempts based on goals based on joining different technologies, such as combining quantization with the latest Transformer model. Methods for implementing NLP on quantum neural networks. In addition to this, there have been a sea of attempts to iterate on conventional techniques through fine-tuning in order to optimize the benefits of existing models, too numerous to list all of them here.

As mentioned in the previous section, the field of NLP has seen another large number of significant developments in recent years. In terms of considering the impact of the results of this paper on the technology, the models in this paper are roughly an offshoot of explaining and modifying this development after the RNN family of methods and before the Transformer model.

 \begin{figure}[htbp]
 	\centering
	 \includegraphics[scale=0.3]{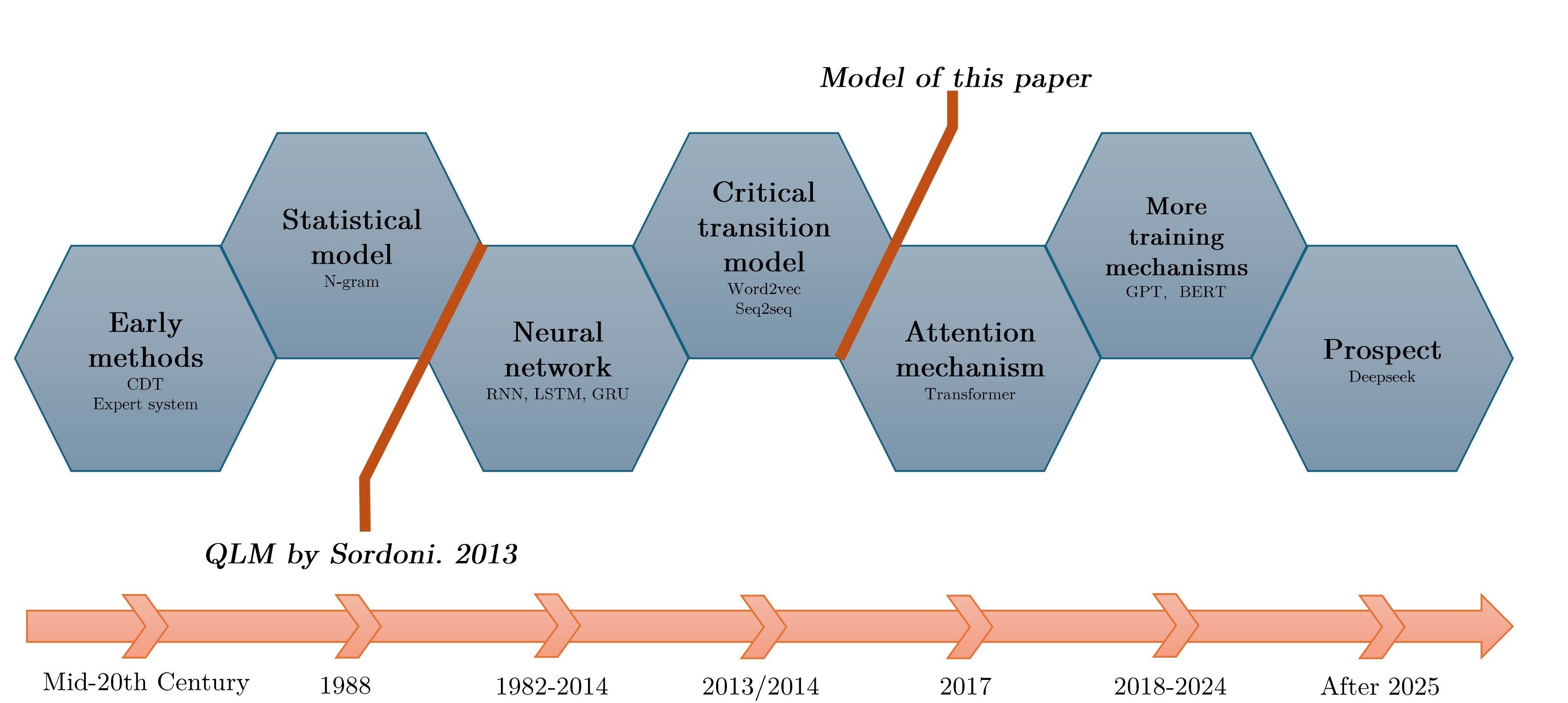}%
	 \caption{NLP development lineage}\label{}
 \end{figure}

\section{III.Intervention of Quantum Theory}

\subsection{3.1 Quantum Language Model}
This section constructs a mapping model for natural language and quantum states. In the study of natural language we are most interested in the meaning represented by a sequence of linguistic or nominal symbols. By choosing an appropriate substrate, the semantics can be represented by a Hilbert space and an algebra on it.
Natural language consists of basic \textbf{linguistic elements} such as words and sentences. We define these linguistic elements mentioned above based on quantum statistical forms. Each linguistic element is regarded as an independent quantum system,\textbf{linguistic element} has the following general form.
\begin{equation}
	\rho_{elem} = \sum_{i}p_{i} \ket{elem}_{i}\bra{elem}_{i} 
\end{equation}

The above definition describes linguistic elements as mixed states, which can be intuitively thought of as being in the quantum state $\ket{elem}_{i}$ with probability $p_{i}$. Such a form allows any linguistic element to be represented as a matrix of finite dimension. Its mathematical form satisfies all the constraints of density matrices.

\begin{CJK}{UTF8}{gkai}
Specifically, the so-called linguistic elements can be character states $\rho_{char}$, word states $\rho_{word}$, sentence states $\rho_{sente}$ or even paragraph states $\rho_{para}$ or even longer natural language arrangements. Or let us step outside of time and space constraints, such as the Chinese character radicals $\rho_{\text{日}}$，All symbols of the entire language family $\rho_{Niger-Congo}$, A summary of all the languages used in the period $\rho_{La\ Troisi\grave{e}me\ R\grave{e}publique}$. \textit{Attention that these linguistic elements may not contain information about serialized structures.} 
\end{CJK}
Density matrices have the advantage of easily describing mixed-state quantum systems, and in many scenarios density matrices provide a more convenient way of dealing with them. However, the fundamental motivation for using density matrices to describe linguistic elements in this paper is based on classical information theory, which states that any linguistic symbol or set of symbols must have information entropy.If linguistic elements are regarded as pure states, then the state of the whole system is completely known, and this can only represent a few special cases that are not general. All natural languages developed by human beings in daily use have uncertainty. This uncertainty is reflected both in the probability amplitude of determining the quantum state of a linguistic element. It is also reflected in the unknown nature of the statistical operators manifesting the specific quantum states of the system. The classical entropy description focuses on the latter. This is a qualitative description, and there is a related discussion of information entropy below.

By \textbf{Hypothesis III}, we assume that the quantum states describing the elements of a language can be unfolded on a substrate consisting of a set of their own properties. A Dirac notation description is used to sidestep the treatment of the representational problem. For the linguistic element $\ket{elem}$ that uses the ket notation constraint, it is a quantum superposition state of the linguistic element. Its state vector can be expanded on a specific attribute base as follows:
\begin{equation}
	\ket{elem} = \sum_{j} c_{j}\ket{attr}_{i} \label{define:elem}
\end{equation}

This means that the linguistic element $\ket{elem}$ is in a superposition state of these substrates $\ket{attr}$. The $\ket{attr}_{i}$ acts as a ground state in quantum mechanics, depending on the chosen substrate. The advantage of this is that in a form analogous to quantum mechanics, quantum states can exhibit different information based on representations.

Natural languages have External Properties, which include lexical properties such as word properties and pronunciation, and Grammatical Properties, which are more complex grammatical concepts such as word forms and word positions. These intrinsic properties can be collectively referred to as extrinsic properties in comparison to the quantum system that describes the elements of the language.

Adopting such a broader definition allows linguists, on the one hand, to study many other extrinsic properties of language at the same time. An interesting property of natural language is that when we study, for example, the various properties of a vocabulary, it still needs to be described in terms of the natural language itself. Indeed, general-purpose language models have demonstrated the ability to solve all types of NLP subtasks together in the form of dialogs. It is hard not to imagine that these properties of language are ``self-describing''. This provides a tricky way to study these extrinsic properties. It makes it always possible to expand on a set of attribute ground states that are spatially complete bases when we represent quantum states.

In this paper, we focus on the property of ``semantics'' that produces results in mainstream NLP methods. In fact, the above description has shown that external properties can also be decomposed into a more basic set of substrates. It can be said that all the above attributes can be represented as some kind of subspace of the ``semantic space''. The property of semantics is essentially an intrinsic property contained in the quantum system.

\subsection{3.2 Semantic Representation}
\subsubsection{3.2.1 Basic Example}
For natural language, the property we are most interested in is the meaning represented by its sequences. We start by building a description of the simpler case and then generalize to a more general theory.

First assume that a linguistic element $w$ has i determined semantics. In this simplest case, we may wish to turn to the dictionary to make the word $w$ have i determined dictionary interpretations. Each of these dictionary interpretations is expressed by some sentence. By \textbf{Hypothesis II} it is understood that these meanings are the meanings that arise from the use of $w$, and which specific meaning should be chosen in the interpretation obviously depends on the contextual environment.

At the same time each lexical meaning of $w$ can be used as a probability amplitude superposition of j underlying meaning fragments. In this example we borrow the linguistic term \textbf{sememe} to refer to these base meaning fragments. A sememe is the smallest unit of meaning that makes up the meaning of a word.

Use $\ket{sema}_i$ to denote the semantic vector of the ith semantic, and a set of semantic vectors $s=\{\ket{seme}_1,\ket{seme}_2 \dots \ket{seme}_j\}$then constitutes the semantic substrate of the word $w$ itself.

Quantum state expansion to density matrices based on its own semantic substrate yields:
\begin{equation}\label{def_elem}
	\begin{aligned}
		\rho_{word} &= \sum_{i} p_{i}\ket{sema}_{i}\bra{sema}_{i} \\
		&= \sum_{n,m}\sigma_{nm}\ket{seme}_{n}\bra{seme}_{m}\\
		&\sigma_{nm}=\sum_{i}p_{i}c^{(i)}_{n}c^{(i)\dagger}_{m}
	\end{aligned}
\end{equation}

Here $c_{j}$ describes the intrinsic probability amplitude of the quantum state, while $p_{i}$ in the aforementioned density matrix describes the statistical properties of the system. The $\sigma$ are the coefficients of the corresponding matrix elements. We can interpret this to mean that each meaning of the linguistic element $w$ is simultaneously in the superposition state of multiple sememes, making it simultaneously have the meanings of these sememes. 

Based on the conventions of quantum mechanics, the set of substrates in the example is named \textbf{eigen semantic substrate}, means that it is  an ``eigenvalue'' of the ``observable'' of semantics. We quantify this in the form of a matrix.

\textbf{Example 3-1}: 

We select the dictionary $ \left\lbrace computer,vector\right\rbrace $ to illustrate.
The word vector appears a total of $40$ times in the publication version of this article, of which $6$ times are used as a surrogate for the meaning of state vectors in quantum mechanics, the remaining $25$ times refer to word embedded in computer technology words, and $9$ appear as pronouns in examples.

Example meanings of the other words selected and their distribution in this paper are shown in the table below:

The individual eigen semantic substrates are represented by mutually orthogonal basis. From the definition (\ref{def_elem}) we compute the matrix representation of each vocabulary as follows.\\
{\noindent}\rule[-7pt]{8.5cm}{0.1em}\\
\begin{equation*}
	\begin{aligned}
		&\rho_{vector}= && \rho_{computer}= &\\
		&\left[ 
		\begin{matrix}
			0.15 & 0 & 0 \\
			0 & 0.625 & 0 \\
			0 & 0 & 0.225
		\end{matrix}
		\right] &&
		\left[ 
		\begin{matrix}
			0.285 & 0 & 0 \\
			0 & 0.25 & 0 \\
			0 & 0 & 0.465
		\end{matrix}
		\right] &\\ 
	\end{aligned}
\end{equation*}
\begin{table}[htbp]
	
	\label{tab:absolute_eval}
	\setlength{\tabcolsep}{4.5pt}
	\renewcommand{\arraystretch}{1.5} 
	\begin{tabular}{l|cccc}
		\toprule
		&$p_{i}$ & $substrate_{v}$ & $p_{i}$ & $substrate_{c}$\\
		\midrule
		$\ket{\text{Sema}}_{1}$ &$\frac{2}{7}$&$\ket{\text{quantum state}}$&$\frac{1}{7}$&$\ket{\text{quantum computer}} $\\
		$\ket{\text{Sema}}_{2}$& $\frac{5}{8}$&$\ket{\text{word mapping}} $& $\frac{1}{4}$  &$ \ket{\text{classical computer}}$   \\
		$\ket{\text{Sema}}_{3}$&$\frac{9}{40}$&$\ket{\text{pronoun}}$&$\frac{13}{28}$&$\ket{\text{pronoun}}$\\
		\bottomrule
	\end{tabular}
	
\end{table}

It should be emphasized that neither the description nor the use of the model in this paper involves the operation of ``observing'' quantum states. Readers familiar with quantum mechanics may easily associate, ``When using (observing) the vocabulary $w$ it interprets (collapses) with probability $c_{i}$ represented as the semantic element $\ket{seme}_i$'' such an interpretation. Such an interpretation seems to be correct in this simple sample. However, we recommend that your majesty do not pay attention to this representation. Priority is given to the mathematical form of quantum states rather than to ``observation'', and emphasis is placed on a particular ground state through the principle of superposition of states in which linguistic elements simultaneously take on the meanings of these elements rather than through certain operations.This is because next, in the application phase of the model, we will manipulate the mathematical form of the linguistic element directly on a classical computer.

\subsubsection{3.2.2 Common Semantic Substrate}

In the previous section we described a general example that is easy to understand, and we can see that both linguistic elements, meanings and semantics in it can be described in some of natural languages. However, when the size of the study object increases, the number of new semantics and substrates that are constantly introduced grows linearly.

We can notice that possibly for different semantics may be able to be constituted by superposition of sememes containing repetitions. 
Then is it possible to find a set of substrates that makes it possible to construct superposition states representing the quantum states of all linguistic elements in the set of objects under study.

Taking this further, it seems that the property that the substrate can be interpreted through natural language is likewise not necessary. At this point the substrate can no longer be strictly called a sememes; it can no longer be described by language. It is clear that we only need to know the formulation of the quantum state of the linguistic elements and ensure that the set of objects under study are all under the same formulation. This is reflected in finding a common set of semantic substrates, from which any operation is formalized.

Also based on quantum mechanical conventions, we might name such a set of substrates for the same set of linguistic elements \textbf{common semantic substrate}.

In the current word embedding techniques used in NLP, it is common to take the mapping of natural language symbols into vectors. This approach stems only from practical feedback and has not been fully grounded in theory. In fact so far we have provided a good explanation combining physical methods and linguistics for this widely used basic technique. And it can be argued that word vectors can be regarded as a simplified approximation of the quantum representation in this paper. Strictly speaking, this is purely accidental in this paper's attempt to quantum modeling language.

\textbf{Example 3-2}:
Let's continue with the previous example.

In this example, the semantics of two vocabularies are further disassembled into basic units containing the same sememes. The density matrix is computed by the equation (\ref{def_elem}) to show the density matrix representation of multiple vocabularies based on the common semantic substrate. \\
{\noindent}\rule[-7pt]{8cm}{0.1em}\\
\begin{equation*}
	\begin{aligned}
		&\text{\textbf{Computer} for meaning \textit{quantum computer}:}\\[0.2em]		
		&\;\;\ket{seme_{computer}}_{1} =  \ket{quantum}+\ket{state}+\ket{machine}\\[0.2em]
		&\text{\textbf{Computer} for meaning \textit{classical computer}:}\\[0.2em]	
		&\;\;\ket{seme_{computer}}_{2} =  \ket{classical}+\ket{state}+\ket{machine}\\[0.2em]
		&\text{\textbf{Vector }for meaning \textit{quantum state}:}\\[0.2em]	
		&\;\;\ket{seme_{vector}}_{1} =  \ket{quantum}+\ket{state}\\[0.2em]
		&\text{\textbf{Vector} for meaning \textit{word embedded}:}\\[0.2em]	
		&\;\;\ket{seme_{vector}}_{2}=  \ket{word}+\ket{mapping}\\[0.2em]
	\end{aligned}
\end{equation*}
\rule{8cm}{0.05em}\\
\begin{equation*}
	\begin{aligned}
		\rho_{vector} =
		\left[ 
		\setlength{\arraycolsep}{5pt}
	\begin{matrix}
	0.083 & 0 & 0.083 & 0 & 0 & 0  \\
	0 & 0 & 0 & 0 & 0 & 0  \\
	0.083 & 0 & 0.083 & 0 & 0 & 0  \\
	0 & 0 & 0 & 0 & 0 & 0  \\
	0 & 0 & 0 & 0 & 0.416 & 0.416  \\
	0 & 0 & 0 & 0 & 0.416 & 0.416  
\end{matrix}
\right] & 
\end{aligned}
\end{equation*}
\begin{equation*}
\begin{aligned}
\rho_{computer} =
\left[ 
\setlength{\arraycolsep}{5pt}
\begin{matrix}

	0.177 & 0.177 & 0.177 & 0 & 0 & 0  \\
	0.177 & 0.333 & 0.333 &0.155 & 0 & 0  \\
	0.177 & 0.333 & 0.333 & 0.155 & 0 & 0  \\
	0 & 0.155 & 0.155 & 0.155 & 0 & 0  \\
	0 & 0 & 0 & 0 & 0 & 0  \\
	0 & 0 & 0 & 0 & 0 & 0  
\end{matrix}
		\right]&
	\end{aligned}
\end{equation*}
\begin{table}[htbp]
	\setlength{\tabcolsep}{10pt}
	\renewcommand{\arraystretch}{1.5} 
	\begin{tabular}{l|cc|cc}
		\toprule
		&\multicolumn{2}{c|}{\textit{vector}}&\multicolumn{2}{c}{\textit{computer}}	\\
		C-S SubStrate& $p_{sema}$& $c_{seme}$ & $p_{sema}$&$c_{seme}$ \\
		\midrule
		$\ket{quantum}$ &$\frac{1}{6}$&$\frac{1}{\sqrt{2}}$&$\frac{8}{15}$&$\frac{1}{\sqrt{3}}$\\
		$\ket{state} $& $\frac{2}{6}$&$\frac{1}{\sqrt{2}}$& $1$&$\frac{1}{\sqrt{3}}$   \\
		$\ket{machine}$&0&0&$1$&$\frac{1}{\sqrt{3}}$\\
		$\ket{classical}$ &0&0&$\frac{7}{15}$&$\frac{1}{\sqrt{3}}$\\
		$\ket{word} $&$\frac{5}{6}$ &$\frac{1}{\sqrt{2}}$&0& 0  \\
		$\ket{mapping}$&$\frac{5}{6}$&$\frac{1}{\sqrt{2}}$&0&0\\
		\bottomrule
	\end{tabular}
	
\end{table}

\subsubsection{3.2.3 Advantages of Quantum State Embedding}

The initial adoption of the word-vector perspective pointed to its attention to the algebraic operations that can be done between the meanings of languages. A commonly cited example is $\overrightarrow{queen} = \overrightarrow{women} + \overrightarrow{king}$, which led researchers to attempt to represent natural language algebraically with vectors.

In mathematical form, quantum pure states can indeed be represented by vectors. The model in this paper points out that the embedding method, treated only as a vector property, does not have various quantum mechanical properties. Within the framework of quantum mechanics, semantic vectors of pure states can only be in a linear superposition of their eigenstates. This provides both new constraints on the mathematical form and new properties of the objects of word embeddings. Performing algebraic operations in vector space extends to operator operations in which Hilbert space satisfies the physical meaning.

As mentioned in Hypothesis II, the meaning of any natural language varies depending on the specific linguistic context. As with the steps to promote substrate in this section, this polysemy is manifested not only in polysemous words, but also in subtle variations of meaning in the context of language use that cannot be rigorously described in natural language. 

For a linguistic element that can be regarded as a mixed state to represent either of its determinate semantics, the subtle semantic changes that arise in a specific linguistic context are reflected in the minor fluctuations of the quantum state. Statistical weighting in the form of density matrices lends itself to the ability to describe both of these polysemous properties.

\begingroup \squeezetable \squeezetable
\begin{table}[!ht]
	
	\begin{ruledtabular}
		\renewcommand{\arraystretch}{1.5} 
		\begin{tabular}{l|ll}
			
			& \textit{Vector }   & \textit{Density matrix}  \\\hline
			
			\textit{Representation}       &   Vector space     & Hilbert Space \\
			\textit{Operation}	         & Mainly basic algebra  & Quantum mechanical operator \\
			\textit{Embedded way}  &     Injective map                & Statistical mixing\\
			\textit{Interpretability}  &    Empirical       & Superposition of sememe-like\\
			\textit{Numerical }  &   Deterministic value       & Allow minor fluctuations 
		
		\end{tabular}
	\end{ruledtabular}
\end{table}
\endgroup
In the next chapter we will also show the many benefits that natural language brings to normalization operations by describing them in the form of density matrices.

\section{IV.INTERACTION OF LINGUISTIC ELEMENTS}

\subsection{4.1 Structured linguistic Elements}

\subsubsection{4.1.1 Linguistic Composition}
Words form sentences and sentences form paragraphs. Natural languages produce longer and more informative structured sequences by means of underlying construction rules. This process is described in this section. The framework of quantum mechanics itself contains a description of the inter-complexity of systems.

For a universal composite of two \textit {not interacting} quantum systems, the state space is the tensor product of the subsystems in general. For a composite of j linguistic elements whose state is at $\rho_{j}$, the state can be expressed as:

\begin{equation}
	\rho^{'} = \rho_{1}\otimes \rho_{2}\otimes \dots \rho_{j} 
\end{equation}

The superscript of the element $\rho^{'}$ resulting from the combination is used to indicate that the base element $\rho$ is of one order higher compared to the former. In this paper, the combined composite quantum system is called \textbf{language composition}. Linguistic composition can also be regarded as a linguistic element.

\begin{figure*}[!htp]
	\centering
	\includegraphics[scale=0.5]{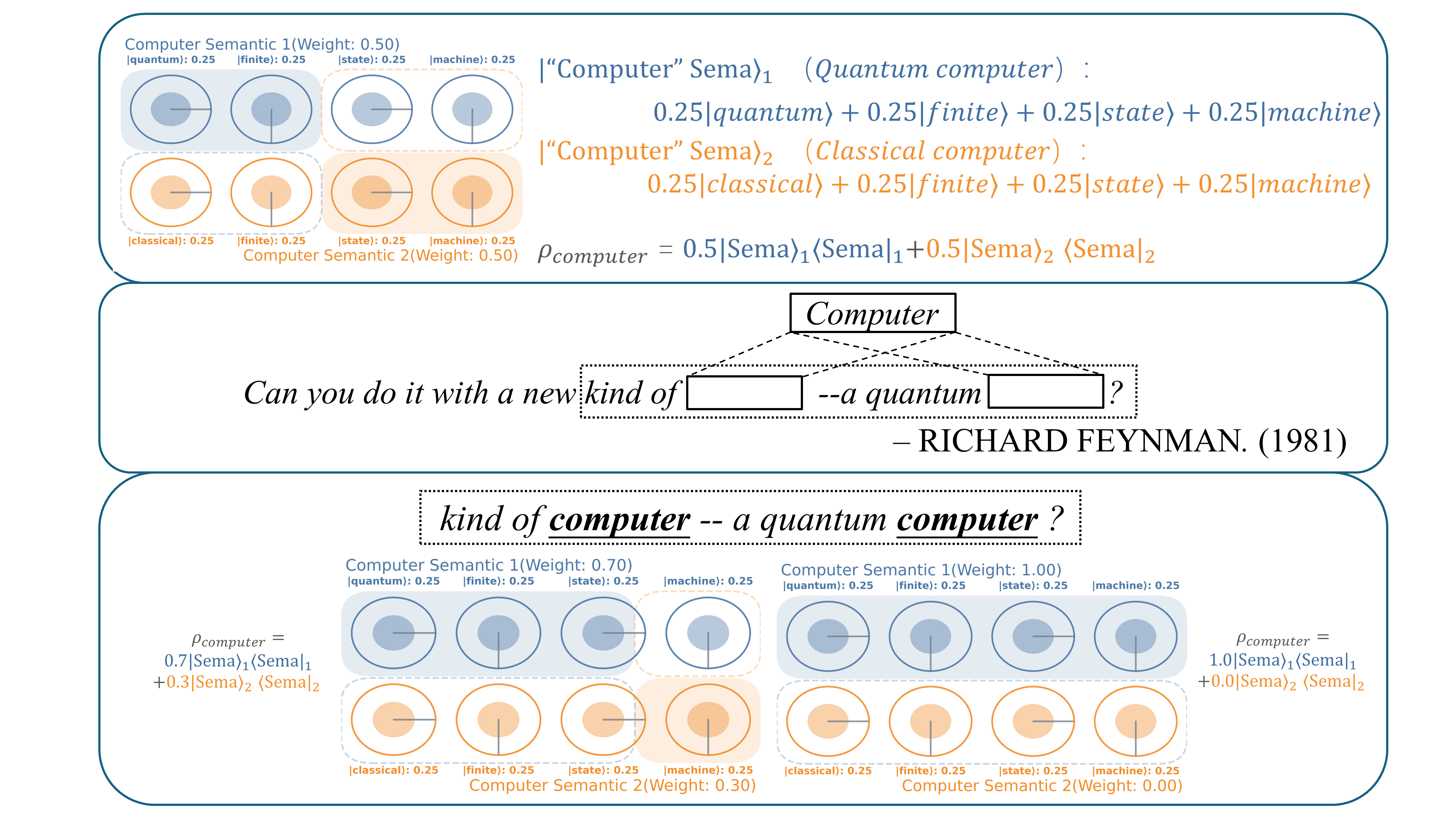}%
	\caption{Example4-1}\label{ex4-1}
\end{figure*}

The composite process can be described in a simplified way as the result of the repetition of the same basic operation. The basic operation for an element to enter the system can be defined as follows:

\begin{equation}\label{metaop}
	\rho^{'}_{end} = \rho^{'}_{start}\otimes \rho_{ins}
\end{equation}

The composite mixed-state matrix can be decomposed into a mixed statistical representation of the underlying pure states of the constituent subsystems:

\begin{equation}
	\rho_{end} = \sum_{i j} p_{i}p_{j}\ket{\phi_i}_{start}\ket{\phi_j}_{ins}\bra{\phi_i}_{start}\bra{\phi_j}_{ins}
\end{equation}
Such a compounding process can be explained quite intuitively in the actual use of language. The meaning most likely to be represented in a sentence comes from the compounding of the most common and dominant semantic of each word.

Obviously such a representation presupposes that the individual elements are considered as independent quantum systems that do not interact with each other. The composite generates a tensor product state whose substrate is the tensor product of the substrates of the elements before the combination. Recall from the examples we have listed above that if the individual elements take on an eigen semantic substrate, the post-composite linguistic composition acquires the semantics of its constituent elements. But for the more general case where we obtain a common semantic substrate for a set of sets of linguistic elements, then compositing the substrates on top of each other on the same set produces some 
idempotent direct product of the substrates.

This seems to imply that higher-order linguistic elements necessarily contain more information as larger quantum systems. It also gives rise to the inference, based on this description, that for a common intrinsic substrate of a finite set of natural language symbols, the meaning of all possible natural language sequences can be represented as a power tensor product of this substrate in its own right. This assumption remains to be verified by linguistic experiments.

\subsubsection{4.1.2 Linguistic Context}
It is important to mention that one of the growing concerns of NLP research in recent years is that natural language is a contextually relevant language. A series of models inherited from earlier times, which considered only the statistical properties of language, did not perform well in new NLP tasks such as generative tasks. It is a technical fact that a qualitative leap in NLP performance has occurred after the introduction of mechanisms that allow contextually ordered information to be encoded into the model.

And the numerical representation of quantum states can adopt this phenomenon as an interplay between quantum systems. During the continuous combination and disassembly of linguistic elements as they enter the context, the state of their semantic space changes in real time with these behaviors, and the information of natural language symbols in the semantic space is presented by statistical aggregation of a single operator through a density matrix.

\textit{This thesis argues that this is precisely the basis for the use of quantum states, noted in the previous chapter, to be able to represent semantic fluctuations of linguistic elements in specific linguistic contexts.}

See FIG. \ref{ex4-1}, Example 4-1 for an illustration of the transformation of the weight of different semantics of a word as it enters a specific contextual setting. The visualization scheme used in the figure uses area to indicate the probability of measuring a particular ground state of a double qbits, with phase not playing a key role in this example but illustrated by the plumb line inside the circle. Each row represents the ground state of a different semantic, and the area in the bottom color of each row illustrates the proportion of the corresponding semantic in the density matrix.

\begin{figure*}[!htp]
	\centering
	\includegraphics[scale=0.5]{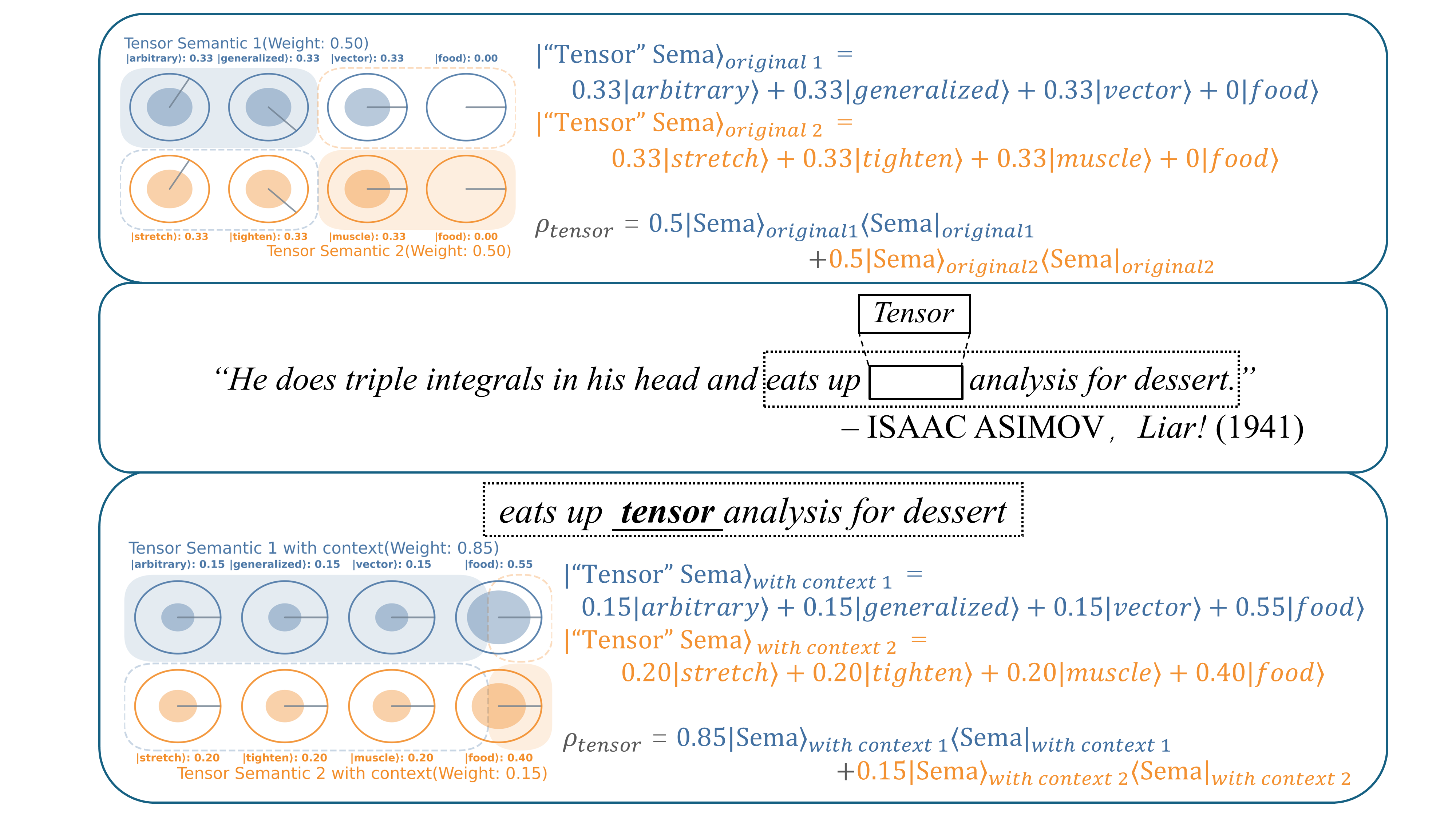}%
	\caption{Example4-2}\label{ex4-2}
\end{figure*}

Example 4-2 of FIG. \ref{ex4-2} represents another phenomenon. The semantic substrate of a language changes when particular words are combined with each other. A substrate that did not exist before appears, or a substrate by itself disappears. The latter indicates that some substrate with a probability amplitude that existed in the subsystem before the composite becomes zero after the composite. This does not follow from the direct product result above, which is called an entangled state in quantum mechanics, often because of interactions between quantum systems.

Returning to language use, this situation corresponds to cumbersome grammatical rules or specific linguistic conventions. These situations prevent the composite process from being described by a simple composite of quantum systems.

\subsubsection{4.1.3 Numerical Approximation of Composition}

In this paper, it is argued that finding a way to specifically represent the density matrix of the post-composite linguistic composition remains an extremely important issue. This is not only about the integrity of the model, but also a crucial step for the model to be applicable. Between the current quantum theory still does not have an exhaustive formal theory of entanglement. This has transformed the problem from a rigorous theoretical one to an engineering one in a limited time. This section briefly points out some directions for consideration and provides some constructive ideas. It also analyzes the applicability of today's fruitful NLP methods as experimental results.

From the current application of NLP methods, designing a composite representation of the word vectors throughout the utterance is also a key step in the models. In the CBOW method simply the average of the word vectors is used as the word vector of the sentence.\cite{word2vec} In a further model from the sequence perspective, a series of neural network architectures called Encoder is designed. This allows the symbols in the sequence to be unified into a result vector in a sequential order with the nonlinear parameters of the network.\cite{seq2seq}

We argue that benefiting from the statistical nature of the density matrix, such a direct averaging method as the former does not destroy the physical meaning of the model. This training method is due to the fact that the composite representation of the word vectors is only used as an intermediate step in training, rather than focusing on the specific meaning of the composite vectors. This direct approach is feasible until a superior method is identified.

And for further sequence models, which take full advantage of the nonlinear fitting ability of neural networks. The manner in which the composite vectors are generated is too opaque to make further discussion. When combined with the content of this paper, we speculate that this composite process may have some theoretical connection to the linguistic ensemble mentioned in the next section. Also this model gives birth to the Transformer model that works well by adding a context window designed to limit the generation of composite vectors, and still deserves further attention.\\

\textbf{Quantum Many-body System }: 
If we continue from the point of view of physical modeling. The serialized nature of natural language makes it possible to treat it as a chain of one-dimensional long quantum systems. For example a quantum many-body system composed of several different atoms, our language model encounters a certain similarity. A method already exists for representing the composite case of such quantum systems called Matrix Product State. which can approximate the density matrix of a one-dimensional quantum chain taking into account phenomena such as entanglement. This paper argues that modeling based on this method would be a strong candidate.

\textbf{Markov Chain }: 
To be precise in this paper the default is a Quantum Markov Process. Markov chains were first proposed in linguistic research, and the development of their self-contained theory has been widely used in NLP tasks. Markovianity expresses the idea that the future state of a system is only related to the current state. We can think of the quantum states of linguistic elements as changing only when the basic operations described above are performed, which makes it possible to study the evolution of the whole system in a discrete way, away from the time variable. Also directly considering the whole language sequence as a Quantum Markov Chain is a generalization that combines classical ideas with the quantum model developed in this paper.

\textbf{Rigorous Quantum Modeling}:
As stated above, the whole sequence can be regarded as a constant repetition of the basic operations to get the result. The object of the study is reduced to the study of small quantum systems into a large environment. Starting from the quantum statistical theory, the change in the state of the small quantum system is obtained by considering the environment as a large thermal reservoir. On the other hand, if we take a closer step to extend the quantum system corresponding to natural language into a quantum mechanical system that strictly conforms to quantum theory. Then after knowing the Hamiltonian quantities of the specific interactions, the effects of the operations on the quantum states of the subsystems and the environment can be described by means of Kraus operators and Partial Trace representations. The next section demonstrates useful conclusions that follow further from this line of thinking.\\
 
The frustration of this section is that obtaining longer structured sequence representations directly from the density matrices of linguistic elements seems to be hampered. We have to do some research on what are tentatively called ``interactions between linguistic elements''. The specifics should be verified experimentally, in addition to theoretical derivations.

\subsection{4.2 Linguistic Evolution}

\subsubsection{4.2.1 Linguistic Ensemble}
Now we want to get something about the macroscopic nature of the elements of language. Let us try to look at the evolution of linguistic elements in terms of quantum statistical physics. A large number of linguistic elements, represented by the same set of substrates, constitute what can be regarded as a kind of \textbf {Linguistic ensemble}.

As the article initially warned about the examples given, the linguistic elements resulting from this combination can also be described by a density matrix. Careful consideration must be given to whether or not the system has structured information within it, that is, whether or not there are interactions. Such a linguistic ensemble synthesis can correspond to a quantum abstract representation of a corpus in linguistics.

For a linguistic element that is not used, e.g. a sentence that is not put into contextual consideration. We consider its state to be static. The state of its next order linguistic element does not change over time. This description also applies to linguistic ensembles, where the description of constraints should be adapted to the case where linguistic elements are not exchanged with other systems for a long period of time. In this case, the states of the linguistic elements within the system do not change over time, and the isolated system is in a thermodynamic equilibrium condition.

The condition for a system that does not exchange energy and particles with the outside world and whose volume remains constant is called a micro-canonical system. If we continue with this analogous quantum theory. Then the conclusions for micro-canonical quantum ensembles can also be applied to linguistic ensembles:

\begin{equation}
	S = -k_{B}\text{Tr}(\rho ln \rho)= -k_{B} ln\Omega (E)
\end{equation}
The entropy $S$ of the system can be derived directly from the density matrix $\rho $ of the system, and $\rho $ is associated with the microstate number $\Omega (E)$ of the system. The $k_{B} $ is the Boltzmann constant. These system functions facilitate the study of the properties of the system of ensembles.

The discussion in the previous section pointed out that the change of state triggered by a linguistic element when it enters the system can be regarded as transient in the study. Independent of time, we discretize the study of the state of the system at each moment. Approximating the system as being in equilibrium allows the entropy of the system to be obtained by applying the above equation.

But let's take the long view. In the practical use of natural language, for example, in the vast number of daily conversations in which various linguistic elements are used, new words are created or old words are given new meanings or old meanings are eroded. As society progresses, new concepts are introduced and people need to find new words to express them. Foreign words enter different languages as cultures spread. The quantum state of linguistic elements changes all the time.

Is there also some kind of pattern or tendency in this evolution? For example, statistical physics assumes that systems evolve to more statistically averaged states, and mechanical systems evolve to energy minima. Does the duality mentioned in the hypothesis I, where symbols are mapped into semantic space, also have some kind of principle as they evolve? To be experimentally verified.

Further, it is necessary to mention the case of information exchange with other systems. This corresponds to a closed system with the condition of a canonical ensemble with only energy exchanges and no particle exchanges with the outside world, and an open system with a grand-canonical ensemble of energy and particle exchanges with the outside world. We note from the above discussion that \textit{energy is here analogized to a quantity of information, while particles are natural language symbols.}

For example, regular conditions can be used to describe attempts to represent the density matrix of linguistic composition in Section 4.1. It is also possible to use canonical or grand-canonical conditions to study linguistic phenomena such as changes in language use brought about by cultural exchanges in a given language family. Rudimentary attempts are made in the applications section of this paper.
\subsubsection{4.2.2 Entropy and Infonergy}

The concept of entropy has long been introduced into linguistics in the study of information theory. In some practical applications that demonstrate its nature, information entropy is used to calculate how many bits are needed to encode an entire character set completely. Or the minimum volume of data compression is calculated by calculating the information entropy to obtain information about the probability of occurrence of information fragments. Here the information entropy is taken as the logarithm of the base 2, which is usually considered to indicate how many bits are needed to represent the number of states of the system under study.

Scholars, including Shannon,\cite{shannon} have applied information entropy to mathematical linguistics, methods that treat languages as Markov chains to study their conditional entropy. The size of the set of symbols in this theory correlates the amount of information contained in the system. This provides some tentative explanation for empirical phenomena such as, for example, the fact that Chinese hieroglyphic texts are usually shorter in length or that Chinese readers read faster.\cite{feng}

Such an interpretive approach actually emphasizes that systematic uncertainty exists in many alternative states, and that a certain number of symbols are needed to represent the full range of possible states. The size of the set of symbols in this theory correlates the information contained in the system. Recalling the duality mentioned in our \textbf{Hypothesis I}, this paper argues that the classical information entropy is a characterization of the number of states of natural language symbols, which concerns the probability of occurrence of these symbols. 

Admittedly it can be further shown that different symbols carry inconsistent amounts of information by comparing the lengths of binary sequences required to show different characters. It can be thought of as a study in progress for the surface symbols in this symbol-meaning duality.

An important inspiration from classical information entropy theory is that \textit{the existence of distinguishable states of a system is a prerequisite for the system to contain information.}

Let's look at an example that points to this specificity. In bioinformatics, information entropy is used to study specific gene expression. Here there is a duality of "gene name (symbol) - gene chain (physical structure)" similar to the symbol-meaning duality. Different gene segments are expressed with different probabilities. When we use information entropy to study gene expression, the information entropy can indicate the probability of expression of each gene segment, but it cannot indicate the information about the number of base pairs on the gene segment. We cannot infer from information entropy that frequently expressed genes have longer gene chains. Obviously the arrangement of base pairs itself carries some information.

This led us to realize that there are also properties that are not included within the notion of the number of microstates of a system, making it impossible to be characterized by entropy. But properties related to the amount of information carried by the system.

Classical Shannon entropy is generalized to quantum information in the form of Von Neumann entropy:
\begin{equation}
	S(\rho) = - \text{Tr}(\rho \log \rho) 
\end{equation}
Mixed-state systems carry uncertainties of their own, and entropy is their statistical property. Von Neumann's form can be related to the entropy of a mixed-state system under the specific conditions of the previous section.

One area where a contradiction arises is that quantum entropy is zero in the pure state. This is because the state of the system in the pure state is completely deterministic in the physical sense although it exhibits uncertainty when observed.

This produced great confusion at first, if we continue to take the classical entropy exposition, it seems that pure states carry no information. But after we have shown that any quantum state in semantic space corresponds to a particular semantics. There is clearly something wrong with a formulation in which a particular semantics contains no information.

Based on these explanatory appearances of contradictions, this paper argues that just as the density matrix describes both the statistical properties of the mixed state and the probability amplitude information of the quantum state, the information can be equally well stored in the probability amplitude inside the quantum state. In this paper, we call this intrinsic property, which mention above, belonging to the representation of quantum states of linguistic symbols as \textbf{Infonergy}.

A prerequisite assumption obviously included here is that each substrate of a natural language quantum state carries the same basic amount of signaling. It is only then that we can quantify the infonergy properties of the two pure states. If we assume more generally that different substrates carry different amounts of infornegy, then it will be impossible for us to make absolute infonergy magnitude comparisons.

We argue that classical information theory only hints at the number of states needed to express all symbols, while a quantum model can further give us access to the amount of information carried by symbols. In statistical distributions expresses information carrying based on the uncertainty of the statistical nature of the system, while in quantum states exhibits a deterministic information carrying based on the quantum nature.

\subsubsection{4.2.3 Physicality of information}

We have cautiously borrowed a large number of quantum theories to explain observed linguistic phenomena. So far, these whimsical theories surprisingly seem to be really feasible.

We wonder why these abstract symbols, born out of the long practice of productive human life, share similarities with physical systems. Is this similarity only in the algebraic structure?

This paper tries to give a fundamental explanation.

First at all is kinds of inference, derived from science itself, that since we cannot observe that the universe exists outside of it, information must exist somewhere within the observable universe. This hypothesis is tentatively referred to in this paper as the \textbf{physicality of information}: information must exist in some arrangement of matter in the universe. This allows humans to use symbols that are surrogates that refer to certain physical structures. We can assume that the information representing language must be stored on top of some quantum-mechanically described system in the universe.

This hypothesis allows us to stop being careful about calling the theory indicated in this paper an ``analogy through the algebraic form of quantum mechanics'' and to try to conform the phenomena of natural language to a strictly quantum mechanical interpretation.

In this interpretive framework, we can boldly state that the infonergy should be analogous to the Hamiltonian. Classical entropy reveals us that different states are the key to assigning meaning to information. In this paper, we argue that the rationality of this approach derives precisely from the principled fact that different states are the basis for discriminating information. Information can likewise be embodied in, for example, discrete energy levels of quantum systems.

For the time being, we can't form any idea about how the substrate of semantic space maps onto a quantum system of discrete energy levels. It is only observed that the two behave similarly. Different energy levels obviously correspond to different energies, making our attempts to quantize infonergy described in the previous section more constrained. 

But what if we go further and stop caring about the size of the infonergy and focus only on the number of states of the infonergy. The informativeness of a language is associated with the size of the Hilbert space of the quantum system.

With this interpretation, although quantum systems may have inconsistent Hamiltonian leading to different structures and ways of evolution. It is still possible to emphasize the focus on the state situation of the system and find a mapping of quantum systems to natural language symbols. And most promisingly. The introduction of analogous Hamiltonian allows us to see ways in which we can completely describe the evolution of natural language corresponding to quantum systems.

The study of the information entropy of language has existed for a long time, but there has been no further study to derive other meaningful ``thermodynamic function'' of language. There is still no conclusive evidence of the meaning of other ``thermodynamic quantities'' in information theory. The characterization of these thermodynamic quantities of information allows us to quickly obtain the state of the linguistic ensemble. These are beyond the original purpose of this paper and will not be further developed.

\section{V.EVALUATION}

\subsection{5.1 Comparative performance with old technology}
Methods for embedding natural language in numerical values for representation have been highly successful without the need for further validation in this paper. This paper provides normative and theoretical support for this technique. The main purpose of this section is to verify that the new method of embedding into density matrices is feasible and to compare the performance of the two on a fair set of tests.

However, there are fundamental differences between the embedded objects. This makes it questionable on example as how to determine that the old and new models have the same parameter scales. Also in the old model there are already many implicit code optimizations and iterations of the algorithm that have been developed over a long period of time in use, which also affects the performance of the results. The results in this section are for reference only. The experimental code is available in the Appendix at the Code Statement, where a more detailed description of the runtime environment is provided.

The current verification of embeddings inevitably uses neural networks. In the problem studied in this paper, it is useful to refer to the process by which humans understand language. All languages are learned later in life to understand how they are used. To a beginner these words can seem blank or mean anything. However, as one is exposed to more and more samples of language use, one eventually forms a correspondence between the symbols and their meanings in the brain. The training of neural networks is an attempt to simulate the above process.

Let us focus on the specific implementation of the model in this paper. One of the difficulties in implementation is that the meaning of the density matrix is much more complex than that of a simple vector, and arbitrarily adjusting the parameters of a vector will only change its properties. While arbitrarily adjusting the parameters of a density matrix may make it no longer legal. Deducing backwards from the results, the density matrix must satisfy several conditions such as hermitian  matrix, semi-positive definite, and the trace is 1.

We use the well-established numerical method cholesky decomposition to keep the data structure satisfying the constraints of the density matrix during training. For simplicity, we use all real numbers in our experiments. Here we also assume that the diagonal elements of the lower triangular matrix are nonnegative and real such that the density matrix corresponds to a unique decomposition matrix.

It is also easy to associate the use of fidelity to replace the cosine similarity method of measuring vector similarity that is needed for training.

The comparison is performed using the CBOW method of embedding training which has been analyzed above and for which the theoretical substitution of quantum states has no effect on the results. The CBOW method is based on the basic assumption that the word vector of a word should be the same as the total vector of the context environment in which it is embedded. Referring also to the paper \cite{eval} for an absolute evaluation of several word embedding methods, we have compared the method of embedding density matrices side-by-side with the embedding method of traditional vectors. The absolute results in each dataset are shown in Table \ref{tab:absolute_eval}.

\begingroup \squeezetable
\begin{table}[htbp]
	
	\caption{Absolute Evaluation Results }
	\label{tab:absolute_eval}
	\setlength{\tabcolsep}{6.5pt}
	\begin{tabular}{lccccc}
		\toprule
		Model & WS & WSS& WSR & MEN & RG65 \\
		\midrule
		Quantum(3Q)   & 0.0382 & 0.0146 & 0.0466  & \textbf{0.0495} & 0.0232 \\
		Classical(36D)& \textbf{0.1160 }& \textbf{0.0638} & \textbf{0.1822}  & 0.0073 & \textbf{0.0651}  \\
		\bottomrule
	\end{tabular}
	\vspace{0.2cm}
	
	\footnotesize 
	
	\textit{Abbreviations:} WS=WordSim, WSS=WordSim-Similarity, WSR=WordSim-Relatedness, RG65=Rubenstein-Goodenough. 
\end{table}
\endgroup
Each dataset is the result of a human assessment of the similarity of word pairs. The evaluation is done by calculating the similarity between the embedded results and the human results, and the Pearson correlation coefficients are tabulated for both. The results of the density matrix embeddings are all positive, and the new model can be seen to have a tendency to learn from the human results.

It should be noted that the density matrix of $8\times 8$ formed by the 3 quantum bits is uniquely determined by its lower triangular matrix in this experimental setting. The number of lower triangular matrix parameters is 36, which is the same number of dimensions as the vector method used for comparison. Rigorously, we can compare the old and new models with the same storage size. Considering the previously mentioned invisible differences, stochastic optimization within the codebase, etc., the comparison here is only qualitatively informative.

The replacement of the embedding target not only redesigns the basis of the word embedding technique, but can also be considered as a theoretical refinement of the vector approach under the theory of this paper. So that there is still a vast room for development in the technology. It is appropriate to focus on verifying the feasibility of the new method in this paper.

\subsection{5.2 Observing Linguistic Evolution}
An important corollary of the second half of the model is that various human linguistic phenomena will have some kind of regularity in their quantum state representation. Between the fact that we did not find a well-formed corpus based on period classification available, this paper attempts to design a simple experiment to demonstrate the feasibility of this approach.

We will analyze a simple corpus consisting of Sino-histories to try to catch the tail of this pattern. There are several advantages of using oriental languages:the Chinese language, although it has undergone changes in its glyphs, has used the same system of symbols since the unification of the script by the First Emperor; and there have been many periods of ethnic fusion and extensive cultural exchanges in the history, which makes it suitable for the above research; the rationale for using the veritable history is that it was written throughout more than a thousand years and the use of the language received punctuation and proofreading in the grammar by the official orthography of the successive dynasties. 

We first train word embeddings by the method of the previous section. And by observing the changes in the existential ensemble function in different periods. Between our lack of knowledge of the role of most thermodynamic functions in language and our lack of a clue as to how to measure infonergy We study first and foremost the von Neumann entropy, a parameter for which a definition can already be given.

If the predictions are met, there will be an influx of new concepts into the language during a period of great ethnic integration such as the Tang Dynasty, which will result in observable changes in some of the coefficients of the Chinese language. In this paper, the period from the Han Dynasty to the Qing Dynasty is divided into nine historical periods, according to the period when each dynasty concentrated its efforts on historical revision.The results of the sub-periods are shown in the Appendix.

It should be briefly noted that it may be pointed out that the Yuan and Ming periods were divided into different periods even though they were written too close to each other. The reason for this division is not only the intuitive division of the dynastic turnover, but also the understanding of the historical situation that there was a tendency to intentionally de-Mongolize the language used in the revision of \textit{The History of the Yuan}. The analysis suggests that such a division helps to reflect the results of this section. For readers concerned with a strict division by period of publication,, also calculated the von Neumann entropy of the four histories of the late Yuan and early Ming as the single set of data:1.6336. For your reference.

The same is true of \textit{Records of Three Kingdoms}, which are categorized as ``'' period, because they are still written in the vernacular language rather than in the vernacular language, and there is a certain degree of inheritance of the language of the Qing dynasty. In short, this paper is not a strict discussion of history. As far as possible, this paper verifies the results on a limited corpus, hoping that readers who emphasize history will bear with us.

The same is true for the classification of the \textit{Records of Three Kingdoms} as a "Han" period, as it still follows the historical writing style created by the Grand Historian, and its use of text is close to that of the Han dynasty history books. In short, this paper is not a strict discussion of history. As far as possible, this paper verifies the results on a limited corpus, hoping that readers who emphasize history will bear with us.

\begin{figure*}[!htbp]
	\centering
	\includegraphics[scale=0.3]{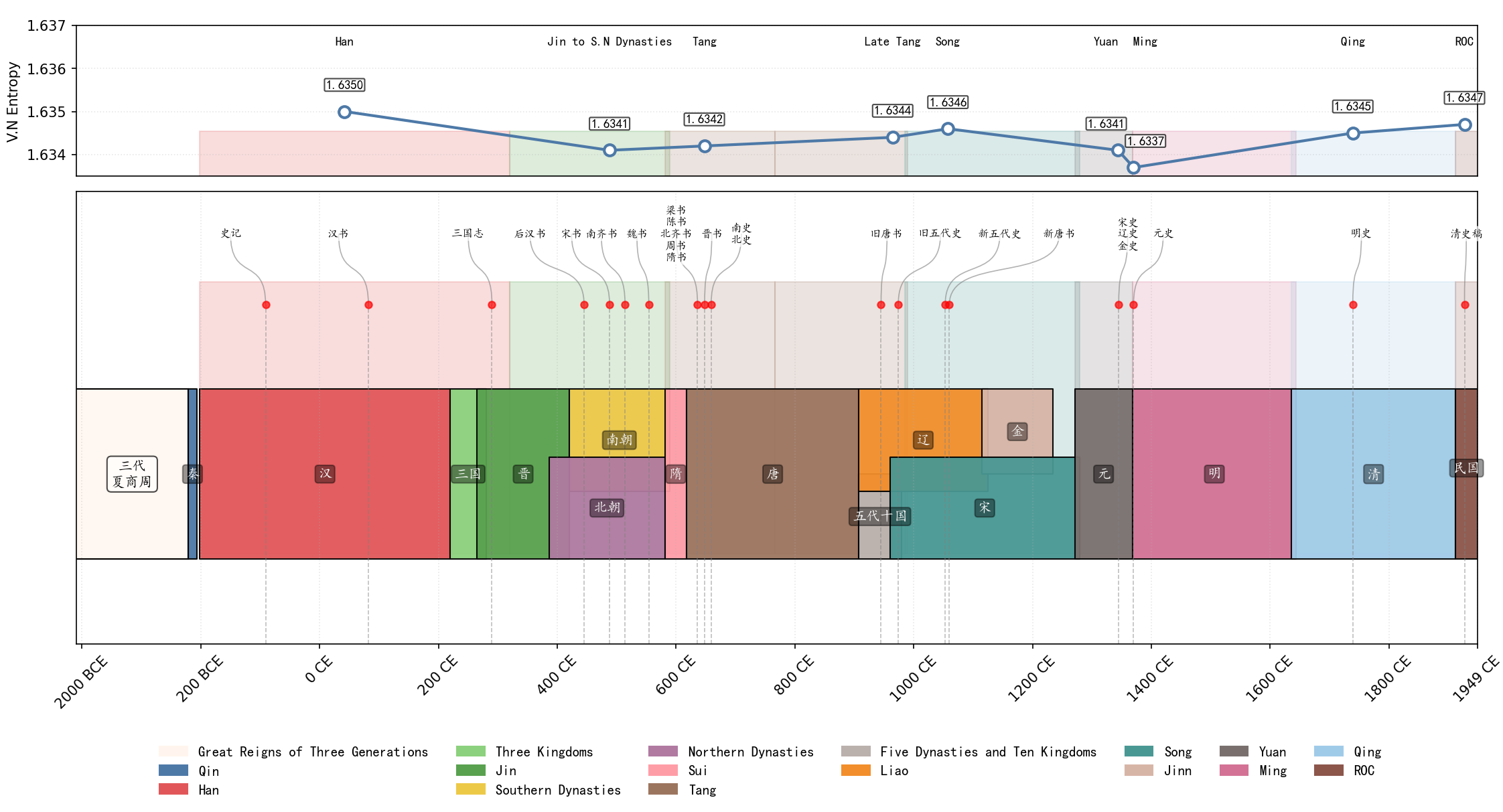}%
	\caption{V.N Entropy of \textit{Histories} of Various Dynasties}
	\label{dyn}
\end{figure*}

The results of the Neumann entropy of the linguistic ensemble at each stage of training, using the model of Conclusion 1 and its parameters, are shown in Fig\ref{dyn}.

Before starting to analyze the results further it is important to note. The authors believe that language, although studied as a mathematical model in this paper, must be clearly understood by the researcher. However, it is important for the researcher to have a clear understanding of the fact that ``language'' in a more general context is linked to many complex social science disciplines.\textbf{ This mathematical approach should only be used as a reference for the study of certain disciplines, such as anthropology.}

First of all there are some areas in this dataset that affect the precision of the experiment. The size of each data varies, and the existing participle programs do not support ancient Chinese well. In addition to this, computer scholars should also consider the influence of various parameters in the training process.

With these aforementioned prerequisites, we can say that in an experiment with limited accuracy,  this paper observes the model speculation in the results. In China, there is a tradition of revising the history of the previous dynasty at the beginning of a new dynasty, and accordingly.For example, this paper considers that the "Late Tang" and "Song" periods, contains the language usage deposits of the Tang dynasty. Accordingly, we have two Von Neumann entropy increases throughout the Tang dynasty and in the period from the Ming and Qing dynasties to the modern era.

The model in this paper states that the decrease in the von Neumann entropy of the linguistic ensemble means that the statistical nature in the language is weakening and the ambiguity of the same word decreases. However, the concepts in human production are always increasing with time, and the amount of information corresponding to the language should also be growing. Accordingly, it can be hypothesized that the amount of infonergy in a language must be increasing over time.

Envisioning real-life language applications for consideration, users of a language exposed to something new tend to borrow existing vocabulary to refer to new concepts, the statistical nature of the language ambiguity increases. And over time, these concepts are normalized with the vocabulary used to refer to them. The ambiguity of the vocabulary gradually decreases again, while the concepts referred to by the language become more complex and infonergy is increasing.

The growth before and after the Tang Dynasty was associated with the integration of peoples within China in the early Tang dynasty and the emergence of a world-connected empire in the middle Tang dynasties. The period from the Ming and Qing dynasties to the modern era was also a time of great technological development and even the modern industrial revolution, which connected all parts of the human world into one. 

Throughout the past two thousand two hundred years, the Von Neumann entropy of language has been on a general downward trend, and as mentioned above, we can see that ``language is digesting concepts''. In this paper, it is argued that information is gradually ``incorporated into the internal state of the quantum system'', and thus the entropy of the ensemble is decreasing.

In the course of this paper, we did try to solve the problems of periodization corpus and ancient Chinese word separation. However, it was soon realized that these workload-intensive projects belonged far beyond the research topic of this paper. The good news is that already from a preliminary result it can be seen that using the theoretical explanation envisioned in this paper has implications for further research. Congratulations!

\subsection{5.3 Universal Quantum Language Model}
\textbf{Classical Computer}: We have already described the implications for existing generalized natural language models. This section addresses the construction of quantum models using the content of this paper and explores the barriers to two construction ideas.

One is the problem of representing composite systems, which has been explored above, and it is clear that the ability to combine and split linguistic elements of arbitrary size is a shortcut to universal. A preliminary discussion of the impact of this paper's model on existing NLP techniques has also been provided in Section 4.1.3. As pointed out in the analysis of the previous chapter, it seems that there is a great difficulty in accurately building the density matrix of linguistic composition from linguistic elements until a sufficient understanding of the entanglement phenomenon is developed. Let's put aside this annoying stumbling block for the time being and see if we can construct some meaningful results for other ideas.

Another route is to integrate quantum state representations into existing NLP technology routes. An example is the attempt to represent linguistic composition under regular systematic conditions mentioned in Section 4.2.1. It cannot be applied because there is still a lack of understanding of key thermodynamic functions such as temperature, partition function, etc. After obtaining a complete description of this transient response, could it be combined with the encoder-decoder model. We speculate that since the model uses neural network brute force to fit the elements to the overall sequence numerical relationships, the model in this paper may already exist in the nonlinear parameters of the network.

And in the technical framework of encoder-decoder, although we generate some kind of representation of the overall sequence in the intermediate steps of the algorithm, it is not the focus of the whole task. \cite{seq2seq}The fact that we do not need to force the overall representation to be a precise representation of the sequence is very effective. Alternatively, the use of quantum state representations allows us to reconfigure the encoder-decoder framework by, for example, simulating a quantum random walk.

\textbf{Quantum computer}: Now let's get back to the ultimate and essential question of quantum engineering, can this model do anything on a quantum computer?

Encoding the density matrix on current quantum circuits is done by separating the pure state and probability terms. This means that the decomposition of the density matrix into a convex combination of $n$ pure states requires a total of $2*log_{2}n$ quantum bits to realize. The $log_{2}n$ principal quantum bits represent the pure states and the $log_{2}n$ auxiliary quantum bits represent the probability terms through a revolving door. Since such a probabilistic superposition always remains pure in the ``quantum data structure'', it always maintains the legitimacy of the density matrix. The calculation of fidelity is also well established using Swap gates.

\begin{figure}[!htbp]
	\centering
	\includegraphics[scale=0.3]{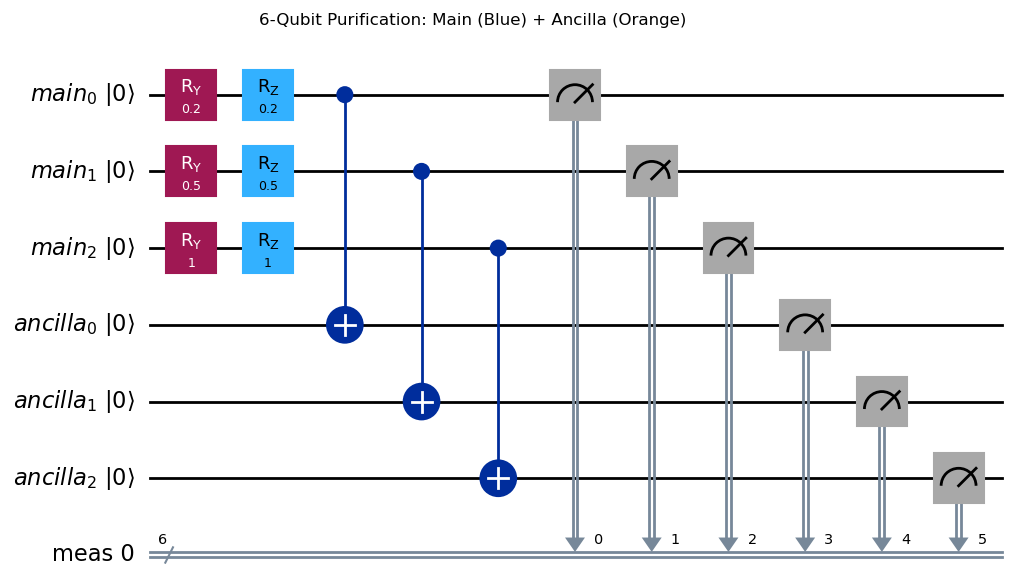}%
	\caption{Quantum circuits for encoding triqubit density matrix}
	\label{qc}
\end{figure}

Both the primary and secondary quantum bits responsible for encoding probabilities grow logarithmically. A schematic representation of a quantum circuit encoding the density matrix of three quantum bits is shown in FIG.\ref{qc}.

Quantum states generated by similar circuits still have continue to be put into training on a neural network designed for a quantum computer. Whether encoding density matrices or promoting neural networks to quantum circuits, such a treatment is accurately described as using quantum circuits to replicate classical algorithms. Looking at it this way it seems to be putting the cart before the horse.

We envision actively making the quantum system in a mixed state, not by splitting its mathematical representation onto a data structure composed of multiple quantum bits. and directly modeling the interactions between linguistic elements mentioned in chapter 3 of this paper through quantum mechanical processes.

Considering a quantum memory of ten quantum bits, we could theoretically encode a 1024-dimensional superposition state. This corresponds to a semantic space of equal dimensionality, which already exceeds even the 768-dimensional vector space currently used in Chatgpt. As far as current technology is concerned quantum computing operations are performed directly on a quantum storage medium, and we refer to this memory for storing quantum linguistic information as Q-drive for short.

If a quantum circuit is designed so that a quantum computer constantly reads (inputs) text (information) while affecting the state of the Q-drive. This allows us to record the quantum states of the linguistic elements directly onto this Q-drive through quantum operations. This allows us to skip the neural network approach, which passes for a brute force reproduction of the system state. This is just like Feynman's reference to simulating quantum phenomena directly through a quantum computer.

This results in a purer and more natural way of obtaining information about quantum states, and is inherently capable of generating new states over time as a result of communication. There is no need to distinguish between pre-trained data sets, etc., or to devise mechanisms to ensure that the data of the language model changes over time or as a result of user interactions.

Obviously, there are natural advantages in generating quantum states, maintaining density matrix legitimacy, fidelity calculations, and other problems Q-drive dictated by the physical architecture.

If we only care about whether the information can be represented, rather than using binary regularization of the information. This even seems to help us build ``Non-Gate-base'' quantum universal computers.

The drawback of doing so is also obvious, based on the axiom that we cannot read the exact state of such a Q-drive. Interactor can only influence its internal quantum state by providing information that allows it to evolve on its own outside of the interactor's recognize. And by interacting with it we get feedback based on its internal quantum state.

It has been said that language is a vehicle for thought. The preceding description makes it seem as if we are interacting with a real intelligence. The feasibility of this line of thinking remains to be seen. At the very least, ``it sounds like we get one step closer to artificial intelligence.''

\section{VI.Conclusion}
The starting point of the model comes from experimental results, namely the remarkable progress in NLP technology in recent years, and a quantum language model is built based on some hypothesis summarized from language use. We review the key research elements of the article:

\textbf{In the first part} The numerical representation of natural language symbols through quantum mechanical algebra is presented and it is pointed out that the existing word embedding technique is a simplified case.

\textbf{In the second part} The process of compounding linguistic elements is discussed. As well as the possibility of studying the macroscopic properties of linguistics and the evolution over time through quantum statistics.

\textbf{In the third section} Applications are discussed, including approximations on classical computers and improvements to existing embedding techniques. The feasibility of using quantum statistics to study natural language is discussed. Specific ideas for future applications are also included.

The abstract meanings of language are in a high-dimensional Hilbert space that is represented by a set of semantic eigen substrate. Any definite meaning can be represented as a quantum state in this space. It is hypothesized that there may exist a common set of eigen substrate that can hold all human languages, such that any semantic meaning can be represented by it.

The basic components of language, in this paper as linguistic elements, are equivalent to symbols that refer to a statistical mixture of superposition states in the space described above. The advantage of this representation is that it can cover small semantic fluctuations in semantics in specific contexts, and cases of multiple meanings at once. Since we do not have physical limitations such as the inability to observe superposition states in reality, we consider the use of linguistic elements to be equivalent to direct access to raw information about language superposition states. (This is of course forbidden under the measurement axiom; we are concerned with its mathematical form.)

We believe that it is possible to study the evolution of language from a quantum statistical point of view based on these models. For isolated language systems, matching micro-canonical ensemble conditions. For other systems with signal and symbol exchanges, other systematic conditions are chosen. For two systems to be complex with each other, we assume that the canonical ensemble condition is followed. It can be hypothesized that some evolutionary processes in which two linguistic ensembles interact with each other can be studied using the external conditions of the grand-canonical ensemble. This is not developed further in this paper.

As to the fundamental explanation for the establishment of this series of models, this paper conjectures that it is based on the principle fact that information exists on physical systems. And accordingly, it is suggested that the energy of the quantum system corresponds to the amount of information system's infonergy to be studied. 

From an application point of view, this paper attempts to provide theoretical modeling of the techniques underlying the large language model, and to fine-tune and improve them accordingly. Attempts are made to design some new computational techniques using principles and algorithms involving quantum. Meanwhile, it is also helpful for further building artificial intelligence on quantum computers.

\section{VII.Concluding Remarks}
The theory of this paper extends interesting queries:
As neuroscientists may ask: Are there quantum processes at the microscopic level of neurons? Is consciousness quantum? We might ask, would the model of this paper be prima facie an indication that concepts of thought are based on quantum structures? Are concepts just a subspace of the Hilbert space of our universe?

On a technical level: does the complex part of a quantum state have any meaning? Are recent popular reasoning language model related to temporal evolution? Is rationality related to the evolution of discretized states?

The study of Rolf Randall is mentioned in the text\cite{Landauer}. Its reading of Laplace's demon problem has, in the past, played an important role in the author's understanding of information theory. Also, we should mention that in the field of linguistics. There have been attempts to find linguistic meaning primitives\cite{an2005}. Although not directly related to the modeling ideas in this paper, this oracles a concerted effort by different disciplines to try to understand human language.

The application of this paper shows us that on the one hand human language contains ideas and on the other hand human language is carried by specific symbols. The semantic space complicates into a thought space. This seems to explain the multimodal capabilities of large language models. We can make the hierarchy: physical structure << information << language << thought. This explanation should provide some support for current technological thinking in artificial intelligence.

In addition, an interesting idea is to form three variables of a language system: symbol (language)--statistical distribution (polysemy)--infonergy (substrate selection), after quantifying the infonergy of natural language symbols. These variables can be balanced on a set of artificially designed linguistic quantum structure devices to design artificial languages.

Symbolic-statistical distribution correspondences can already be formulated for classical information theory. Examples include the space-for-time trick of hieroglyphics and epigrams, where the former exchanged a larger character table for a shorter sequence length. If we base it on all three, or only adjust the symbol-infonergy relationship to weed out linguistic polysemy. Making natural language like a mathematical language draws on the fact that mathematical symbols are highly meaning-specific. To construct a \textbf{harmonize language}, which make the symbols-infonergy reach a mathematical optimization optimal solution. Individual symbols of this harmonize language will correspond to the unique pure state of the system.

Quantum mechanics continues to influence our perception of nature in unexpected areas. Thanks to the mathematical language of linear algebra, otherwise distant fields of study exhibit similar mathematical structures. This is the intention of the term ``intervention'' in the title of this paper.

\clearpage
\newpage

\appendix*

\begin{CJK}{UTF8}{gkai}

\begin{table*}[htbp]
\setlength{\tabcolsep}{11pt}
\renewcommand{\arraystretch}{1.08}
\caption{Explanation of the period classification of \textit{The Twenty-Five Histories} Corpus  }
\begin{tabular}{lllll}
\toprule
\textbf{Period} & \textbf{Title} & \textbf{Author(s)} & \textbf{Publication era} & \textbf{Dynasty}  \\
\midrule
\multirow{3}{*}[-6ex]{\rotatebox[origin=c]{90}{\textbf{Han}} } 
& \textit{史记} &  司马迁 & 征和二年 & 西汉  \\
& \textit{Record of the Grand Historian} & Sima Qian & 91 BCE & Western Han  \\ 
\addlinespace

& \textit{汉书} & 班固等 & 建初五年 & 东汉  \\
& \textit{Book of Han} & Ban Gu, et al. & 80 CE & Eastern Han  \\ 
\addlinespace
& \textit{三国志} & 陈寿 & 太康年间 & 西晋  \\
& \textit{Records of Three Kingdoms} & Chen Shou & 280-290 CE & Western Jin  \\ \addlinespace
\midrule
	\multirow{4}{*}[-1.7ex]{\rotatebox[origin=c]{90}{\textbf{Jin to S.N Dynasties}} } 
& \textit{后汉书} & 范晔 & 元嘉九年至二十二年 & 刘宋  \\
& \textit{Book of Later Han} & Fan Ye & 432-445 CE  & Liu Song  \\ \addlinespace

& \textit{宋书} & 沈约 & 永明五年至六年 & 梁  \\
& \textit{Book of Song} & Shen Yue & 487-488 CE & Liang  \\ \addlinespace

& \textit{南齐书} & 萧子显 & 天监十三年至大同三年间 & 梁  \\
& \textit{Book of Northern Qi} & Xiao Zixian & 514–537 CE & Liang  \\ \addlinespace

& \textit{魏书} & 魏收 & 天保二年至五年 & 北齐 \\
& \textit{Book of Wei} & Wei Shou & 551–554 CE & Northern Qi  \\ \addlinespace

	\midrule
	
	\multirow{8}{*}[-16ex]{\rotatebox[origin=c]{90}{\textbf{Tang}} } 
& \textit{晋书} & 房玄龄等 & 贞观二十年至二十二年 & 唐\\
& \textit{Book of Jin} & Fang Xuanling, et al. & 646-648 CE & Tang \\ \addlinespace

& \textit{梁书} & 姚思廉 & 贞观三年至十年 & 唐  \\
& \textit{Book of Liang} & Yao Silian & 629–636 CE & Tang  \\ \addlinespace

& \textit{陈书} & 姚思廉 &贞观三年至十年 & 唐 \\
& \textit{Book of Chen} & Yao Silian & 629–636 CE & Tang \\ \addlinespace

& \textit{北齐书} & 李百药 & 贞观三年至十年 & 唐  \\
& \textit{Book of Northern Qi} & Li Baiyao & 629–636 CE& Tang  \\ \addlinespace

& \textit{周书} & 令狐德棻等 & 贞观三年至十年 & 唐  \\
& \textit{Book of Zhou} & Linghu Defen, et al. & 629–636 CE & Tang  \\ \addlinespace

& \textit{隋书} & 魏徵等 & 贞观十年\&显庆元年 & 唐  \\
& \textit{Book of Sui} & Wei Zheng, et al. & 636 CE \&656 CE $^{*}$  & Tang \\ \addlinespace

& \textit{南史} & 李延寿 & 显庆四年 & 唐 \\
& \textit{History of Southern Dynasties} & Li Yanshou & 659 CE & Tang  \\ \addlinespace

& \textit{北史} & 李延寿 & 显庆四年 & 唐  \\
& \textit{History of Northern Dynasties} & Li Yanshou & 659 CE & Tang \\ \addlinespace
	
	\midrule
	
	\multirow{2}{*}[-1ex]{\rotatebox[origin=c]{90}{\textbf{Late Tang}} } 
& \textit{旧唐书} & 刘昫等 & 开运二年 & 后晋  \\
& \textit{Old Book of Tang} & Liu Xu, et al.                                                                                                                                                                                                                                                                                                                                                                                                                                                                                                                                                                                                                                                                                                                                                                                                                                                                                                                                                                                                                                                                                                                                                                                                                                                                                                                                                                                                                                                                                                                                                                                                                                                                                                                                                                                                                                                                                                                                                                              & 945 CE & Later Jin \\ \addlinespace

& \textit{旧五代史} & 薛居正等 & 开宝七年 & 北宋  \\
& \textit{Old History of Five Dynasties} & Xue Juzheng, et al. & 974 CE & N. Song  \\ \addlinespace
	
	\midrule
	\multirow{2}{*}[-3.5ex]{\rotatebox[origin=c]{90}{\textbf{Song}} } 
& \textit{新五代史} & 欧阳修 & 皇祐五年 & 北宋  \\
& \textit{New History of Five Dynasties} & Ouyang Xiu & 1053 CE & N. Song  \\ \addlinespace

& \textit{新唐书} & 欧阳修等 & 嘉祐五年 & 北宋  \\
& \textit{New Book of Tang} & Ouyang Xiu, et al. & 1060 CE & N. Song  \\ \addlinespace
	\midrule
	\multirow{3}{*}[-4.3ex]{\rotatebox[origin=c]{90}{\textbf{Yuan}} } 
& \textit{辽史} & 脱脱等 & 至正四年 & 元  \\
& \textit{History of Liao} & Toqto'a, et al. & 1344 CE & Yuan \\ \addlinespace

& \textit{金史} & 脱脱等 & 至正四年 & 元 \\
& \textit{History of Jin} & Toqto'a, et al. & 1344 CE  & Yuan  \\ \addlinespace

& \textit{宋史} & 脱脱等 & 至正五年 & 元 \\
& \textit{History of Song} & Toqto'a, et al. & 1345 CE & Yuan  \\ \addlinespace

	\midrule
\end{tabular}

\end{table*}
\newpage
\begin{table*}
\setlength{\tabcolsep}{20pt}
\renewcommand{\arraystretch}{1.10}
\caption*{Explanation of the period classification of \textit{The Twenty-Five Histories} Corpus(continuation sheet)  }
\begin{tabular}{lllll}
	\toprule
	\textbf{Period} & \textbf{Title} & \textbf{Author(s)} & \textbf{Publication era} & \textbf{Dynasty}  \\
	\midrule
		\multirow{1}{*}[-0.5ex]{\textbf{Ming}}  
	& \textit{元史} & 宋濂等 & 洪武三年 & 明  \\
	& \textit{History of Yuan} & Song Lian, et al. & 1370 CE & Ming \\ \addlinespace
		
		\midrule
		
		\multirow{1}{*}[-0.5ex]{\textbf{Qing}} 
	& \textit{明史} & 张廷玉等 & 乾隆四年 & 清\\
	& \textit{History of Ming} & Zhang Tingyu, et al. & 1739 & Qing \\ \addlinespace

		\midrule
		
		\multirow{1}{*}[-0.5ex]{\textbf{ROC}}
	& \textit{清史稿} & 赵尔巽等 & 民国十六年 & 民国 \\
	& \textit{Draft History of Qing} & Zhao Erxun, et al. & 1927 CE & ROC  \\ 
		
		\bottomrule	
\end{tabular}

\end{table*}

\end{CJK}


\end{document}